\pdfoutput=1 

\documentclass[11pt]{article}

\usepackage{EMNLP2023}
\usepackage{times}
\usepackage{latexsym}
\usepackage[frozencache=true,cachedir=.]{minted} 
\usepackage{booktabs}

\usepackage[T1]{fontenc}

\usepackage[utf8]{inputenc}

\usepackage{microtype}

\usepackage{inconsolata}

\usepackage{tipa}

\usepackage{graphicx}

\title{Molyé: A Corpus-based Approach to Language Contact in Colonial France }

\author{Rasul Dent, Juliette Janès, Thibault Clérice, Pedro Ortiz Suarez, Benoît Sagot \\ Inria, Paris \{firstname.lastname\}@inria.fr}

\begin{document}
\maketitle
\begin{abstract}
Whether or not several Creole languages which developed during the early modern period can be considered genetic descendants of European languages has been the subject of intense debate. This is in large part due to the absence of evidence of intermediate forms. This work introduces a new open corpus, the Molyé corpus, which combines stereotypical representations of three kinds of language variation in Europe with early attestations of French-based Creole languages across a period of 400 years. It is intended to facilitate future research on the continuity between contact situations in Europe and Creolophone (former) colonies.
\end{abstract}

\section{Introduction}
Between the 15th and 19th centuries, several languages developed in colonized territories, which, while sharing a large amount of vocabulary with existing European languages, differ considerably in morphology and syntax. These languages are often labeled English-based [or lexified] Creoles, French-based Creoles, Portuguese-based Creoles, etc., according to the language they share most of their vocabulary with, which is itself called the lexifier. One long standing question has been why the grammars of these languages diverged from their lexifiers to a greater extent than the vocabulary \citep{sousaHistoryCreoleStudies2019}. Much of the difficulty in answering this question stems from harsh social conditions discouraging linguistic documentation and environmental conditions destroying much of what had been documented \citep{mcwhorterCreoleDebate2018}.  

For French-based Creole languages (FBCLs), which developed on islands and isolated continental settlements during the 17th and 18th centuries \cite{chaudensonCreolizationLanguageCulture2001} \footnote{Except Tayo in 19th century New Caledonia.}, reliable documentation largely dates from the mid-late 18th century onward \citep{hazael-massieuxTextesAnciensCreole2008}. However, we note that the formative period of FBCLs coincided with a period of French political and cultural dominance, accompanied by extensive literary production. The cultural works of the period are replete with numerous stereotypes of the speech of several social groups, such as the Germanic Baragouin of Swiss soldiers which, despite various issues detailed by \citet{ayres-bennettVoicesSourcesSeventeenthCentury2000}, demonstrate some specific morphosyntactic features characteristic of FBCLs.

Here, we introduce the Molyé corpus, which regroups 68 works containing literary representations of such stereotypes and/or early attestations of FBCLs, curated from a larger collection of 255 documents identified at the time of publication.\footnote{The corpus can be accessed and downloaded at the following address: \url{https://github.com/defi-colaf/Molye}.} We begin by giving an overview of related corpora and how we approach historical linguistics as an instance of multi-label language identification. After giving some linguistic context, we also explain the process of identifying Creole-like features in French literary works, encoding said works into XML-TEI, and then compiling groups of quotes into a timeline. Finally, we present summary statistics and conclude by giving examples of how our corpus highlights intra-European language contact.

\section{Related Work}
In recent years, Creole languages have garnered attention in the field of natural language processing. On the one hand, \citet{lentWhatCreoleWants2022} have explored how these languages challenge the assumed desirability of certain applications. On the other hand, \citet{lentAncestortoCreoleTransferNot2022} and \citet{robinsonAfricanSubstratesRather2023} argue that language models for concrete problems may shed light on theoretical issues as well. Simultaneously, in computational historical linguistics, \citet{listOpenProblemsComputational2024} has declared the inferral of morpheme boundaries and the detection of contact layers to be major open problems. In this work, we address the paucity of early Creole documentation and the issue of multiple layers of language contact through the applied lens of language identification.

\subsection{(Digital) Diachronic Corpora}

For several regions, such as Louisiana \citep{neumann-holzschuhTextesAnciensCreole1987}, the Caribbean \citep{hazael-massieuxTextesAnciensCreole2008}, Réunion \citep{chaudensonTextesCreolesAnciens1981}, and Mauritius \citep{bakerCorpusMauritianCreole2007, chaudensonTextesCreolesAnciens1981}, diachronic corpora have been compiled in print. However, to our knowledge, only the Mauritian corpus has been systematically digitized and made readily accessible \citep{fonsingCorpusTextesAnciens}. Beyond this, certain historical works have been digitized for inclusion in analysis-oriented private diachronic corpora \citep{mayeuxRethinkingDecreolizationLanguage2019}, or for applied goals like machine translation \citep{robinsonKreyolMTBuildingMT2024a}, and others have been digitized individually by groups such as the ``Groupe Européen de Recherches en Langues Créoles'' \citep{hazael-massieuxCreolicaRevueGroupe2013, hazael-massieuxGroupeEuropeenRecherches2013}. 


To digitize documents in a way that can facilitate reuse, we rely on the standards of Text Encoding Initiative (TEI) \citep{teiconsortiumeds.TEIP5Guidelines2023}. Adherence to these guidelines has produced diachronic corpora which span several centuries, such as \citet{favaroCreationDiachronicCorpus2022}. For languages of France, \citet{bermudezsabelSettingBilingualComparable2022} have addressed some of the challenges of building comparable  corpora for the parent-daughter pair of Latin and French. Similarly, \citet{ruizfaboCreationCorpusFAIR2020a} explore how digitizing 19th century Alsatian theatre aids sociolinguistic studies.

\subsection{Multi-Label Language Identification}
\label{LID}

Algorithms for determining the language of a given text generally rely on tokenizing the text and comparing the tokens against a learned (or explicitly defined) representation of a language \citep{jauhiainenAutomaticLanguageIdentification2019}. For analytic languages written in the Latin alphabet (i.e. FBCLs), tokens generally align with either words or letters. For closely-related languages, there is sometimes only a difference of a singular word or expression between one variety and another, even in longer documents \citep{ljubesicLanguageIndentificationHow2007, caswellLanguageIDWild2020}. In these cases, we can specify disjunctive features such as words/phrases that are thought to separate the varieties to either affirm or reject a label. In the absence of such features, the same string may be valid in multiple languages, which can make it more accurate to assign multiple language labels to the same string \citep{bernier-colborneDialectVariantIdentification2023, kelegArabicDialectIdentification2023}.

\section{Linguistic Background}
\label{Linguistic Background}

The backbone of our corpus is applying multi-label language identification based on disjunctive features across time. In concrete terms, we sought out distinctly ``Creole'' features in Europe before and during the colonial expansion. As such, we briefly review a few characteristics of FBCLs, followed by French literary stereotypes.

\subsection{French-based Creole Languages}

\subsubsection{Description}
\label{Creole Background}
 
While the notion that all Creoles can be defined in purely linguistic terms, as explored by \citet{mcwhorterIdentifyingCreolePrototype1998, bakkerCreolesAreTypologically2011}, is controversial, FBCLs are agreed to share several traits which distinguish them from standard French. Firstly, they generalized the use of tonic pronouns in places where the former would use weak clitic pronouns \citep{syeaFrenchCreolesComprehensive2017}. In cases where French does not have a weak pronoun (i.e. ``nous''), they still differ by not allowing preverbal cliticization of object pronouns. Additionally, while French relies on a system of fusional conjugations, where verb endings mark person, number, tense, aspect and in the case of the past participle, gender, at the same time, FBCLs add person-invariant combinations of Tense-Aspect-Mood (TAM) markers \citep{syeaFrenchCreolesComprehensive2017, bakerIsleFranceCreole1982}. These differences are demonstrated by the anteriority marker ``té'', and the conditional marker ``sré'' in the phrase ``Pour sûr si vou \textbf{té capab changé} vou lapo pou so kenne, vou \textbf{sré pa di} non'' \citep{mercierHabitationSaintYbarsOu1881}. Furthermore, FBCLs do not have an explicit copula in several structures where one is required in French (and English), as demonstrated by the phrases ``Comme \textbf{vous bel}'' (how \textbf{you} [are] \textbf{beautiful}) and ``vou papa riche'' (you[r] \textbf{dad} [is] \textbf{rich}) in Figure \ref{fig:prose}.


\subsubsection{Theories of Origins}

As previously stated, the relationship Creole languages to lexifiers remains a topic of intense debate. For this work, one relevant hypothesis, as explored by \citet{chaudensonCreolizationLanguageCulture2001}, suggests that the accumulation of the defining characteristics occurred over several waves of second language acquisition, as opposed to being the result of a complete break in transmission of syntax, as suggested by \citet{mcwhorterCreoleDebate2018} and \citet{thomasonLanguageContactCreolization1988}. Another line of inquiry explores the extent to which ``foreigner talk'', which is to say a particular kind of simplified register that people adopt when they feel their interlocutors do not have sufficient competence in the language, may have contributed to certain developments in Creole morphology and syntax \citep{fergusonForeignerTalkName1981, fergusonCharacterizationEnglishForeigner1975}. For Portuguese- and Spanish-based Creoles, there is a long history of triangulating Iberian versions of foreigner talk with early modern literary stereotypes and contemporary Afro-Hispanic varieties to get an idea of the range of linguistic variation in the early modern Iberian empires \citep{kihmLinguaPretoLanguage2018, lipskiBozalSpanishRestructuring2001}. In the following section, we explore how a similar approach can applied to French. 

\subsection{French Literary Stereotypes}
\label{Stereotypes}

Up to the 20th century, most people in France spoke regional languages \citep{lodgeFrenchDialectStandard2003}. In the Northern half of mainland France, most of these languages are part of the Oïl dialect continuum, which is itself part of a larger Western Romance continuum. However, non-Romance languages such as Breton (North-West) and Flemish (North) are spoken as well. From the Middle Ages on a particular Oïl variety, associated with prestigious actors was gradually codified into the standard language of the Kingdom of France. This variety was also adopted as a \textit{lingua franca} throughout Europe,  as an alternative to Latin.  During the 17th and 18th centuries, the process of codification culminated in a well delimited variety known as Classical French. %

However, the codified ``bon usage'', was not the only supralocal speech used in France. Even within the Paris region, there was a great deal of variation within what could be considered  ``French'' \citep{wittmannGrammaireCompareeVarietes1995}. In broad terms, we distinguish three types of variation: dialectal and sociolectal variation from the Oïl domain, standard French with regional accents, and interlanguages, especially from L1 speakers of Germanic languages \footnote{Other phenomena, such as the mix of various forms of Occitan in \textit{Monsieur de Pourceaugnac} described by \citet{sauzetLucetteMonsieurPourceaugnac2015}, are beyond our immediate scope.}. In all three of these cases, we find stereotyped combinations of a finite number of highly stigmatized features in a variety of works, including plays, novels, songs, and personal letters. 


\subsubsection{Peasant French}
By the early 1600s, several features of rural usage in the outskirts of Paris (and Western France), such as the combination of clitic pronoun ``je'' with the plural affix ``-ons'', were developed into a convention for representing lower class characters in literature \citep{lodgeMolierePeasantsNorms1991, ayres-bennettSociolinguisticVariationSeventeenthcentury2004}, as seen in this example from  \textit{La Mère confidente} \citep{marivauxMereConfidente1735}: ``\textbf{Je} sav\textbf{ons} bian ce que c'est; \textbf{j'ons} la pareille.''  Although this stereotype was frozen relatively early on, the highlighted combination was used in France and its colonies throughout the colonial period and still exists in Acadian French in particular, albeit more commonly as a plural form \citep{kingFirstpersonPluralPrince2004}. 

\subsubsection{Gascon Accent}
French also came to be spoken as a second language in areas where the regional languages were even more different from French. In these case, the native languages had some influence on pronunciation. In classical French theatre, one common stereotype of such regional pronunciation is the Gascon accent, which can be identified through its betacism (conflating b and v) and fronting of the schwa (replacing e with é). The character Fontignac from \textit{L'île de la raison} \citep{marivauxIleRaison1727}
 demonstrates the convention with this line: ``...\textbf{b}ous m\textbf{é} d\textbf{é}mandez c\textbf{é} qu\textbf{é} \textbf{b}ous êtes 
    ; mais j\textbf{é} n\textbf{é} \textbf{b}ous \textbf{b}ois pas ;
    mettez-\textbf{b}ous dans un microscope.''

\subsubsection{Germanic Baragouin}
\label{Baragouin}
Germanic Baragouin \footnote{The word ``baragouin'' [gibberish])  was also used to describe a variety of contact phenomena ranging from accented pronunciation to genuine pidgins like that used with the Caribs in the Lesser Antilles \citep{wylieOriginsLesserAntillean1995}.}  (henceforth just Baragouin) is our name for a group of stereotypes which simultaneously combine traits of foreigner talk, foreign accents, and Oïl dia- and sociolectal variation. In the early modern period, there are two main variations: the Anglo-Baragouin attributed to English (and Scots) speakers,  and Continental Baragouin associated with German and Dutch, and more specifically, Swiss and Flemish speakers \citep{leachRippingRomanceRibbons2020, dammDeutschfranzosischeJargonSchonen1911}.  A third, industrial-era Flemish Baragouin also developed around the turn of the 20th century in the cities of Tourcoing and Roubaix near the French-Belgian border \citep{landreciesConfigurationInediteTriangulaire2001a}. The main differences between these sub-groups of Baragouin lie in phonetics. The Continental Baragouin generalizes final-consonant devoicing into a complete neutralization of several consonant pairs, such as /b/-/p/, /k/-/g/, /v/-/f/ and /t/-/d/. Similarly, the industrial-era Flemish version features palatal fronting of /\textipa{S}/ and /\textipa{Z}/ to /s/ and /z/. These traits are mostly absent in the English version. 

In terms of morphosyntax, Baragouin shares some traits with Creoles, such as the generalization of strong pronouns, weakening of grammatical gender, and reduced verbal inflection. However, Baragouin also retains an overt copula and systematically inserts third-person pronouns before verbs, which results in sentences such as ``\textbf{Toi \textit{li} être}, par m\textit{on} foi, la plus \textit{p}elle meilleure h\textit{i}meur du monde'' \citep{guelettePremiereParade1714}. The latter features have a special importance, which we explore further in Section~\ref{Missing Li}. 



    

\section{Corpus Creation}
\label{Corpus Creation}
The compilation of the corpus was realized in three overlapping phases. During the first phase, we identified documents which contained n-grams {} thought to be highly disjunctive between French and various FBCLs. After identifying the documents the next step was to convert them relevant samples into the XML-TEI schema of a broader project. Lastly, we classified the documents by location and period and extracted the relevant quotes into a combined XML document to facilitate the preliminary analysis presented in Section~\ref{Preliminary}.

\subsection{Document identification}
The basic strategy was to search Gallica, the digitized library of the Bibliothèque Nationale Française, Delpher, its Dutch equivalent, and later Google Books for disjunctive n-grams. Examples include monograms, (e.g. ``mo'', ``to'', ``yé'') bigrams, e.g (``mo(n) femme'',``mo(n) z'enfant''), and higher n-grams. Due to variation in both French orthography and the conventions/contact varieties themselves, an iterative approach was taken, with documents collected on the first pass providing more ``unusual'' n-grams for subsequent searches. In the earliest stages, we did not note the exact searches, but later began to record the search terms as well. In a later stage, we also added several Creole sources known through secondary literature in order to facilitate in-depth diachronic comparison.

Because we are working with stereotypes, a certain level of similarity was to be expected.  Nevertheless, in some cases, we found that certain works go into the realm of explicit reference and/or pastiche. For direct quotation, there is \textit{Les fêtes de l'amour et de Bacchus} which includes a reprise of the linguistic humor from \textit{Le Bourgeois Gentilhomme}, among other pieces. As far as pastiche, we can highlight the early 16th century \textit{Testament du Gentil Cossoys} and its early 17th century reprise, the \textit{Testament d'vn Escossois}. The latter is a simultaneously condensed and updated version of the former. Thus where the original reads ``Adiou par tout nobe royaulm de Frans /
Adiou comman le povre pals de Cos...'' \citep{smithTestamentGentilCossoys1920}, the reprisal has `` Ady par tout le Royaume de France/
    Premierement ady le pay de Coss...'' \citep{desigogneTestamentVnEscossois1620}

\begin{table}[h]
    \centering\small
    \begin{tabular}{lrrr}
    \toprule
        Search & Lang Type & Document & Year\\
        \midrule
        ``ly va'' & Baragouin & Francion & 1630 \\
         li-même & Peasant  & L'Épreuve & 1740 \\    
         conné li & L. Creole & L'autre monde &  1855\\
         \bottomrule
    \end{tabular}
    \caption{Sample Searches and Documents}
    \label{tab:my_label}
\end{table}

\subsection{Encoding Documents}
Given both the large number of documents it was necessary to establish an order of priority for incorporating works into the corpus. We initially focused on both Baragouin and Peasant French in works of classical theatre that had already encoded by sources such as theatre-classique.fr  \citep{fievreTheatreClassique2007}. Beyond the core of classical French theatre, however, a wide variety of genres are represented. These include poetry, songs, religious material, short prose, and an entire novel. The subject matter exhibits a similar degree of variability. In the Baragouin section alone, we find, among other things, two mock-testaments, a criticism of military leadership, a love letter, and a discussion about the political implications of an ongoing civil war. 

After treating the extant XML, we explored semi-automatic generation of XML-TEI documents from semi-structured sources such as Wikisource, as well as directly from scanned documents. In the former case, we used relatively simple custom Python scripts to facilitate conversion to TEI, such as wrapping all of the lines in a \texttt{‹p›} (paragraph) or \texttt{‹l›} (line/verse) tag, and then identifying divisions and headers manually.  In the the latter case, this involved a considerable amount of manual transcription due to the diversity of genres and formats. For shorter works, such as poems and songs, we used eScriptorium \citep{kiesslingEScriptoriumOpenSource2019a} to perform text recognition with the CATMuS Print model \citep{gabayReconnaissanceEcrituresDans2024}. However, more complex layout (e.g. newspaper) were transcribed manually. For longer works, we entered the relevant quotes directly into a file of excerpts.
 
\subsection{Linguistic Annotation}
Since this corpus is in large part intended to illustrate a sociolinguistic continuum assigning discrete linguistic labels poses distinct challenges. Although it is clearly anachronistic to speak of ``[Colony] French/Creole'' before the founding of a given colony, we observe that in certain cases, namely in Réunion and Louisiana, the ``approximative French'', ``pidginized French'', or ``pre-Creole'' (depending on one's point of view) bears striking continuity with Baragouin at the morphological and syntactic levels. In a parallel fashion, early texts which are clearly ``Creole'', such as ``La passion de Notre Seigneur selon St Jean en Langage Negre'', display combinations of features which make it difficult to say \textit{which} Creole based on purely linguistic data. 

Following the brief outline given in \ref{Linguistic Background}, we distinguish between five main kinds of language: Classical French (met-fr), Peasant French (fra-dia), (Gascon) Accented French (fra-gsc), Baragouin (subdivided into fra-ang, fra-deu, and fra-nld), and (pre)-Creoles. The Creole portion is in turn subdivided into four regions and labelled using the respective ISO codes: Réunion (rcf), Louisiana (lou), Haitian (hat), and French Guianese (gcr). For the initial work, we have somewhat simplified the question of diachronic and dialectal continua by assigning one label based on the territory a document claims (or has been presumed) to represent, with the exception of grouping the earlier ``Flemish'' baragouin with the German one rather than the later Flemish Baragouin, based on the differences described in Section~\ref{Baragouin}.

\begin{figure}[!htp]
\begin{minted}[tabsize=1, fontsize=\scriptsize, xleftmargin=5pt, xrightmargin=5pt]{xml}
<div type="scene" n="10">
    ...
    <sp who="JACQUES" xml:lang="mau">
        <speaker>JACQUES.</speaker>
        <p>... Enfin pourtant , li jetté son zépée , 
            li remetté pistolet dans son place , 
            li prendre son plume , li assisé tranquille ,
            et li fini écrire sa billlet là moi porté vous.
            Ah vlà li. 
        </p>
    </sp>
    <sp who="STRAFFORD" xml:lang="fra-ang">
        <speaker>STRAFFORD lit le billet haut.</speaker>
        <p>» Vous avez raison , monsieur , 
            je suis mort pour vous et pour votre ami » . 
            <stage> ( Il parle. )</stage> 
            Toi voir lui mort  [etc...]
        </p>
    </sp>
    <sp who="BELTON" xml:lang="met-fr">
        <speaker>BELTON.</speaker>
        <p>Moins que jamais ; 
            c'est absolument une énigme pour moi.</p>
    </sp>
</div>
\end{minted}
    \vspace*{-5mm}
    \caption{This excerpt from Scene 10 of \textit{Le duel singulier} \citep{dorvignyDuelSingulierComedie1800} shows how we tag language usage by speaker. It includes standard French alongside  Anglo-Baragouin and an unspecified Creole with Mauritian characteristics. [formatting adjusted] }
    \label{fig:drama_tei}
\end{figure}

\begin{table*}[t]
    \centering\small
    \begin{tabular}{llrrl}
    \toprule
    \multicolumn{1}{c}{Target/Region} & \multicolumn{1}{c}{Label} & Works & Tokens & \multicolumn{1}{c}{Timespan}\\
    \midrule
    Normative French & met-fr & 35 & 37066 & 1649-1779 \\
    Peasant & fra-dia & 14  & 27825 & 1665-1740 \\
    Gascon & fra-gsc  & 4 & 4530 & 1672-1800 \\
    Anglophone & fra-ang & 4 & 4441 & 1509-1800\\
    Continental Germanic & fra-deu & 25 & 6899 & 1580s$\sim$1779 \\
    Flemish (Tourcoing/Lille) & fra-nld & 4 & 2664 & 1880-1932 \\
    Réunion & rcf & 3 & 10713 & 1760s, 1830s\\
    Lesser Antilles (Martinique) & gcf & 2 & 477 & 1671 \\
    Haiti & hat & 4 & 7395 & 1730s$\sim$1802\\
    Louisiana & lou & 10 & 26068 & 1748-1895\\
    French Guiana & gcr & 2 & 43414 & 1796, 1885\\
    Mauritius (tentative) & mau & 1 & 196 & 1800 \\
    \bottomrule
    \end{tabular}
    \caption{An overview of the linguistic and temporal spread of the corpus.}
    \label{tab:overview}
\end{table*}

For adding linguistic labels to documents, we used two complementary rule-based strategies. For plays where one character (or more) uses non-standard speech throughout, we simply identified the \texttt{‹sp›} (speech) tags associated with that character and inserted an \texttt{xml:lang} attribute with the corresponding label, which allowed us to keep associations between characters and speech turns. Additionally, we added tags at the \texttt{‹p›} level to facilitate text extraction.

For prose, keeping track of specific characters was more difficult. Initially, we tried implementing key-ngram-based regex patterns. Because our languages of interest are frequently embedded in longer French passages, a preprocessing step of sentence tokenization was implemented. Although our disjunctive n-grams generally correspond to words, we use character-level regex patterns that incorporate a special boundary symbol to minimize multi-level tokenization. For the initial annotation, the presence of any one disjunctive n-gram was sufficient to trigger the relevant label. While this method was very useful for highlighting interesting passages, manual retouching was necessary to fix issues of imperfect sentence tokenization, as well as missed examples. In Figure \ref{fig:prose}, we find a reported clause in Louisiana Creole that is not marked because it contains no disjunctive words, followed by a reporting clause in French (``dit l’esclave d’une voix caressante''),  that is unintentionally included with correctly identified  Creole speech in the following sentence. The third sentence is marked as expected.

\begin{figure}[!htp]
\begin{minted}[tabsize=1, fontsize=\scriptsize, xleftmargin=5pt, xrightmargin=5pt]{xml}
<p>
    « Comme vous bel !
    <s xml:lang="lou"> dit l’esclave d’une voix caressante ;
    vou gagnin ain ti lair si tan comifo ! </s> 
    <s xml:lang="lou">vou popa riche, mo sûr ;
    di li achté moin.</s>
    ...
</p>
\end{minted}
    \vspace*{-5mm}
    \caption{Uncorrected semi-automatic annotation of \textit{L'Habitation Saint-Ybars} \citep{mercierHabitationSaintYbarsOu1881}}
    \label{fig:prose}
\end{figure}

\subsection{Compiling Extracts}
After adding language tags at the document level, we created a composite timeline that balances facilitating direct comparison between excerpts with giving some level of contextualization. For plays, we extracted scenes where at least one of the \texttt{‹sp›} turns contained an \texttt{xml:lang} attribute with an appropriate value, as demonstrated by Figure \ref{fig:drama_tei}. By extracting the entire scene, we include samples of normative French and retain the coherence of the conversation to some extent. For monolingual poems, we included the entire poem, albeit possibly excluding meta-linguistic commentary such as notes. For prose,  we implemented a multi-level extra process of first trying to identify broad tags  like \texttt{‹p›} based on the \texttt{xml:lang} attribute, and then narrower tags like \texttt{‹s›} only if they were not already included as part of a broader group.  In Figure~\ref{fig:prose}, the overall paragraph would be assumed to be French, so only the lines within the \texttt{‹s›} tags would be extracted, which is why correcting the linguistic annotation is important.

\subsection{Balancing}
As exemplified by the Gascon accent, the literary conventions can be summarized using a relatively short list of rules. This means that there is a degree of diminishing returns to adding additional examples once we have a basic understanding of said rules. As such, we did not concern ourselves with attempting to create a statistically balanced corpus. In particular, due to the more labor-intensive nature of (semi)-manual encoding, we deprioritized the Peasant French variety early on, which has already received more careful study, and instead focused on the earliest and latest attestations of Baragouin. This may create the impression that literary Peasant French was primarily a 17th century phenomenon. However, this stereotype remained in use until the 19th century. Along similar lines, we did not include many attestations of Mauritian Creole precisely because a digitally accessible diachronic corpus to the same effect already exists \citep{fonsingCorpusTextesAnciens, bakerCorpusMauritianCreole2007}.

\section{Corpus Presentation}
Overall, we found 255 historical works which demonstrate features relevant for the history of FBCLs. We have selected excerpts from 68 of these works to form the basis of the first version of the corpus. The earliest text is the ``Testamen d'un Gentil Cossois'', written anonymously around 1509, and the most recent is Jules Watteuw's ``Belle Réponse'', published in 1932. The main corpus consists of a single, publicly available XML file containing bibliographic information for the collection, followed by a body which contains ``TEI'' tags that regroup the relevant selections from each work and are accompanied by their own brief bibliography section. From this file, one can create customized subcorpora that correspond to specific questions by specifying a date range and the language labels that are to be considered.





\begin{table*}[t]
    \centering\small
    \begin{tabular}{lrrrrr}
    \toprule
    \multicolumn{1}{c}{Target/Region} & Infinitive & Inflected & TAM & CE & Tokens\\
    \midrule
    Normative & 105 & \textbf{1328}  & 129 & 254 & 37066\\
    Peasant & 76 & \textbf{1006} & 129 & 251 & 27825\\
    Gascon & 14 & \textbf{131}  & 16 & 47 & 4530\\
    \midrule
    Anglophone & \textbf{74} &32 &7  &5 & 4441\\
    Continental Germanic & \textbf{89} & 62  & 11  & 13 &6899\\
    Industrial Flemish  & 0 & \textbf{44} & 0 & 18 & 2664\\
    \midrule
    Réunion & 5 & \textbf{125} &54 &2 & 10713\\
    \midrule
    Haiti & 0 & 157 & \textbf{102} & 27 & 7395\\
    Louisiana & 10 & 1086  & \textbf{944} & 129 & 26068\\
    French Guiana & 1 & 1001 & \textbf{950} & 40 & 43414\\
    \bottomrule
    \end{tabular}
    \caption{Attestations of different forms of ``être''. TAM and CE cover creolized inflection.}
    \label{tab:etre}
\end{table*}

At present, the corpus contains a total of 188,866 tokens (whitespace tokenization), excluding metadata. Because of the historical focus of the text, all of the primary sources are in the public domain, and most are readily consultable online. In these cases, we also retain cached copies with additional bibliographic information. In the cases where quotes have been included from printed secondary sources, we do not include metalinguistic commentary. Table \ref{tab:overview} provides a high-level summary of the varieties we distinguish and their relative sizes and time spans.

\section{Preliminary Results}
\label{Preliminary}
Since the main effort of this work has consisted of gathering and grouping multiple non-standardized varieties, proceeding directly to quantitative methods presents special challenges. For the initial demonstration, we provide a few qualitative observations and show how we can support them through relatively simple frequency-based methods, with a particular focus on the relevance of Baragouin \footnote{The following section uses broad IPA in bold.}.

 \subsection{First Person Pronoun: Mo(è)}
 During the colonial era, the French pronoun ``moi'' had two primary variants : \textbf{mwe} and \textbf{mwa}. FBCLs can be grouped according to which form of ``moi'' became the subject pronoun. The first group, consisting of Haitian and Lesser Antillean Creoles, predominantly uses \textbf{\textipa{mw\~E}}, which is clearly a nasalized version of \textbf{mwe} \citep {hullOriginChronologyFrenchbased1979}. The second group, comprised of Mauritian, Seychellois, French Guianese, and Louisiana Creoles, uses the form \textbf{mo}. This division corresponds to further differences in the pronominal system, with the first group also using case-invariant pronouns and marking possession through postposition, while the second group distinguishes between subject and oblique variants and uses proposed possessive adjectives \footnote{Exceptionally, Réunion uses \textbf{\textipa{mw\~E}} with case distinctions.}. Although \textbf{mo} is tied to \textbf{mwa}, its exact origins are less clear. Furthermore, there is documentation that \textbf{mo} was once used by the first group, before being replaced in the 1900s \citep{hazael-massieuxTextesAnciensCreole2008}. 

Several of our documents shed new light on the relationship between these two variants. Firstly, beyond the canonical \textbf{mo}, we also found examples of ``moué,``moé'', ``moè'', ``moë'', and ``moa'' in 19th century Louisiana alone. In  \citet[p.189]{jobeyAmourNegre1860}, for example, includes ``\textit{Moè} té cré bien, \textit{moè} perdu papier la yest'', which combines the Caribbean-like \textbf{mwe} with the Louisiana-specific definite plural marker \textbf{laje} (spelled ``la yest''). By itself, this can be explained by 19th century New Orleans' status as a crossroads of French- and Creole-speaking networks. Secondly, however, we found numerous attestations of \textbf{mo} in Flemish Baragouin. For example, the opening line of ``Poutche'' \citep{watteuwPitche1927} is ``Accoute un fos, \textbf{mo} ne pas bête''. The latter may help explain \textbf{mo} as one innovation which diffused from Europe alongisde {mwe}, rather than a parallel innovation.

\subsection{Copula: ê(tre)}
Additionally, we noticed that Baragouin has a tendency to overuse the infinitive  ``être'' (to be), rather than either conjugating the verb like French, or omitting the copula as in FBCLs. We began quantifying this variation by measuring the frequency of two basic patterns: the infinitive, and all inflected forms. We further tracked two subsets of inflected forms forms that have been integrated into various FBCLs: (precursors of) TMA markers (\textbf{(e)te)}, \textbf{s(r)e}', \textbf{s(r)a} and orthographic variants thereof), as well fusions involving the pronoun ``ce''. For demonstrative purposes, we set aside the samples for the Lesser Antilles and Mauritius, since they  are particularly limited. Unfortunately, we could not take into account the clause-restricted copula \textbf{je} due to it being homophonous with the much more frequent third-person plural pronoun and a derived plural marker in Louisiana and French Guiana.

 Table \ref{tab:etre} demonstrates the results of this experiment. As expected, Normative French, Peasant French, and Gascon-accented French all use a wide variety of inflections. In contrast, the FBCLs the Americas retain specific grammaticalized  uses, such that ``être'' is rare, while inflected forms largely correspond to either TAM markers or presentatives with ``ce'' \footnote{And \textbf{je} which we left out as explained above.}. Réunion, which is distinguished among FBCLs for retaining French auxiliaries, stands out as transitional. In contrast to both groups, both Anglophone and Continental Baragouin  (but notably not later Flemish) generalize use infinitive ``être'' more than inflected forms, but do not completely discard the latter.

\section{Discussion}
\subsection{Missing (L)(i)nks}
\label{Missing Li}
By itself, the generalization of ``être'' shows that decreased use of inflection and copula deletion, two traits of  FBCLs suggested to indicate pidgin origins by \citet{mcwhorterCreoleDebate2018}, did not \textit{necessarily} develop at the same time nor for the same reason.  Beyond this, however, we are able to directly tie one process underlying the generation of Baragouin to one Creole language in particular: Réunion Creole (RC). 

As \citet[p.393]{hullTransmissionCreoleLanguages1993} observes, the subject pronoun \textbf{li}, shared by all FBCLs, is employed by a Swiss German in \textit{Le Bourgeois gentilhomme} in place of ``il''. More specifically, as \citet{dammDeutschfranzosischeJargonSchonen1911} remarks, the systematic insertion of this third-person pronoun before verbs, as mentioned in Section~\ref{Baragouin}, is particularly reminiscent of RC, where we find sentences such as ``\textbf{Moi \textit{i} crois} vrai, bien vrai dans mon cœur n’en a bon Dieu'' \citep{bolleeDeuxTextesReligieux2007}. In both Baragouin and RC, this preverbal pronoun also fuses with auxiliaries, as in this example from \textit{Les filles errantes} \citep{regnardFillesErrantes1690}: ``Moi \textbf{l'être} un étrangir qui cherchir à logir dans sti ville.'' and the Réunionese ``Moi \textbf{l'est} bien content voir à vous'' \citep{heryFablesCreolesExplorations1883}. 

 Although the exact function and source of the preverbal marker in Réunionese Creole are both debated, one common interpretation is that it marks finiteness on verbs and originated as a generalization of third person reprise pronouns \citep{bolleeDeuxTextesReligieux2007}. Interestingly, a similar generalization of third-person \textbf{object} pronouns  is observed in Spanish-language representations of Africans as early as the 17th century, and comparable phenomena continue to exist in varieties of Spanish in the Americas influenced by Quechua and Nahuatl \citep{lipskiBozalSpanishRestructuring2001}. In our corpus, we also observe that ``li'' in particular also appears in Peasant French, primarily as a clitic indirect object. As Baragouin also inserts preverbal pronouns in sentences that use the French first-person subject clitic ``je'', the inserted preverbal pronoun corresponds to a few homophonous French subject, object, and adverbial pronouns. This in turn suggests our corpus is  relevant for contact scenarios beyond FBCLs.
\subsection{The Bigger Picture}

Beyond tracking individual features, our corpus offers a window into the broader sociolinguistic context of French in the early modern period. In the case of the first person pronoun, despite the temporal mismatch, the specificity of ``mo'' points to the Low Countries  as a point of interest. Upon closer examination, several works spell out a network connecting Swiss soldiers to this region and Paris  in the context of the French-Hapsburg wars such as a 1692 ``Air suisse ou flamand'' which references the Nine Years' War in Mons, Namur and Maastricht directly, This detail is of interest for Louisiana and Mauritius, where German-speaking settlers and soldiers played important roles in the French colonization in the 1720s. \citep{vaughanCreatingCreoleIsland2005, klingler2003if, bakerIsleFranceCreole1982}. 

 Along similar lines, ``Le duel singulier'' stands out as a ready-made case study. This play combines  normative French, the Gascon accent, Anglo-Baragouin, and an unspecified Creole, as exemplified in Figure \ref{fig:drama_tei}.  As such, it bolsters theories that the FBCLs of the Caribbean region may developed during the period of Anglo-French cooperation during the early 17th century on islands such as Saint-Christophe and Tortuga \citep{parkvallRoleStKitts1995}. Furthermore, the Baragouin can be cross-referenced against the  Law French of English courts of that period \citep{lofstedtNotesBeginningsLaw2014}.

\section{Conclusion}
In short, we have introduced the Molyé corpus, a new resource which puts French literary stereotypes alongside early forms of several French-based Creole languages. We have shown that restructuring of the French pronominal and verbal systems are attested throughout the 16th, 17th, and 18th centuries, and specifically associated with speakers of Germanic languages. Although stereotypes like the conventionalized Baragouin only address a fraction of the real linguistic variation of the period, our corpus nevertheless raises important questions about how people communicated in lands where French and Germanic languages came into contact. Furthermore, it shows that at least some  of the divergences between FBCLs and French can be traced back to developments which were already underway in Europe.


\section*{Limitations}
The major constraint of this work has been converting unstructured works into XML-TEI. As mentioned in the methodology, this involved complete re-transcription in some cases. Overall, we found more than 200 pertinent documents, but were only able to include one third of them. In particular, we had to leave out works in regional languages of France such as Picard, Walloon, and Poitevin. Similarly, we did not address some relevant phenomena, such as the 17th century Carib Baragouin and the 19th century Tirailleur French in order to maintain the scope of the work.  Although we are well aware of such varieties, we found few instances using our method, and thus leave them as natural targets for future work.

\section*{Ethics Statement}
The main idea of this article is that European literary stereotypes from before and during the colonial period can help fill in the some gaps in the early history of (French-based) Creole languages. As such, many of the primary and secondary sources that we have compiled contain negative imagery and commentary regarding various social groups. Sharing such sources should not be taken as endorsement of the views contained therein.

\section*{Acknowledgements}

This work was primarily funded by the Inria ``Défi''-type project COLaF. This work was also partly funded  by the last author's chair in the PRAIRIE institute funded by the French national agency ANR as part of the ``Investissements d'avenir'' programme under the reference ANR-19-P3IA-0001.

\bibliography{anthology,molye}
\bibliographystyle{acl_natbib}

\appendix



\end{document}